# Building Bridges: Viewing Active Learning from the Multi-Armed Bandit Lens

**Ravi Ganti and Alexander G. Gray**
Computational Science and Engineering.
Georgia Institute of Technology
Atlanta, GA 30332
gmravi2003@gatech.edu, agray@cc.gatech.edu

## Abstract

In this paper we propose a multi-armed bandit inspired, pool based active learning algorithm for the problem of binary classification. By carefully constructing an analogy between active learning and multi-armed bandits, we utilize ideas such as lower confidence bounds, and self-concordant regularization from the multi-armed bandit literature to design our proposed algorithm. Our algorithm is a sequential algorithm, which in each round assigns a sampling distribution on the pool, samples one point from this distribution, and queries the oracle for the label of this sampled point. The design of this sampling distribution is also inspired by the analogy between active learning and multi-armed bandits. We show how to derive lower confidence bounds required by our algorithm. Experimental comparisons to previously proposed active learning algorithms show superior performance on some standard UCI data-sets.

## 1 Introduction

In the classical passive binary classification problem one has access to labeled samples $\mathcal{S} = \{(x_1, y_1), \ldots, (x_n, y_n)\}$, drawn from an unknown distribution $P$ defined on a domain $\mathcal{X} \times \{-1, +1\}$, where $\mathcal{X} \subset \mathbb{R}^d$. The points $\{x_1, \ldots, x_n\}$ are sampled i.i.d. from the marginal distribution $P_\mathcal{X}$, and the labels $y_1, \ldots, y_n$ are sampled from the conditional distribution $P_{Y|X=x}$. Classical learning algorithms, such as boosting, SVMs, logistic regression choose a hypothesis class $\mathcal{H}$, and an appropriate loss function $L(\cdot)$, and solve some sort of an empirical risk minimization problem to return a hypothesis $\hat{h} \in \mathcal{H}$, whose risk $R(h) \stackrel{\text{def}}{=} \mathbb{E}_{x,y \sim P} L(yh(x))$ is small. However, in domains such as speech recognition, natural language processing, there is generally a lacuna of labeled data, and obtaining labels for unlabeled data is both tedious, and expensive. In such cases it is of both theoretical and practical interest to design learning algorithms, which need only a few labeled examples for training, and also guarantee good performance on unseen data. In recent years active learning (AL) has emerged as a very popular framework for solving machine learning problems with limited labeled data (Settles, 2009). In this framework the learning algorithm is "active", and is allowed to query, an oracle $\mathcal{O}$, for the label of those points which it feels are maximally informative for the learning process. The hope is that by using few, but wisely chosen labels the active learning algorithm will be able to learn as well as a passive learning algorithm, which has access to lots of labeled data.

Broadly speaking AL algorithms can be classified into three kinds, namely membership query (MQ) based algorithms, stream based algorithms and pool based algorithms. All these three kinds of AL algorithms query $\mathcal{O}$ for the label of the point, but differ from each other in the nature of the queries. In MQ based algorithms the active learner can query $\mathcal{O}$, for the label of a point in the input space $\mathcal{X}$. However, this query need not necessarily be from the support of the marginal distribution $P_\mathcal{X}$. MQ algorithms might work poorly when the oracle $\mathcal{O}$ is a human annotator (Baum and Lang, 1992). Stream based AL algorithms (Cohn et al., 1994; Chu et al., 2011) sample a point $x$ from the marginal distribution $P_\mathcal{X}$, and decide on-the-fly whether to query $\mathcal{O}$ for the label of $x$. Stream based AL algorithms are computationally efficient, and most appropriate when the underlying distribution changes with time. In pool based AL we are provided with a pool $\mathcal{P} = \{x_1, \ldots x_n\}$ of unlabeled points, which have been sampled i.i.d from the marginal distribution $P_\mathcal{X}$, and a labeling oracle $\mathcal{O}$, which when queried for the label of $x$, returns $y \sim P_{Y|X=x}$. Algorithms in the pool based setting have the luxury of deciding which points to query by looking at the entire pool.

**Contributions.** In this paper we shall deal with pool based active learning. We shall also assume that we have a query budget, $B$, of the maximum number of queries the AL algorithm can issue to the oracle.

*We view AL from the lens of exploration-exploitation trade-off.* The concept of exploration-exploitation is central to problems in decision making under uncertainty, and is best illustrated by the multi-armed bandit (MAB) problem [1] [2]. The MAB problem is a $B$ round game, where in a generic round $t$, the player has to pull one among $k$ arms of a multi-armed bandit. On doing so the player suffers a loss $L_t$. The player does not get to know the loss he could have suffered if he had pulled a different arm. The goal of the player is to minimize the cumulative loss suffered over $B$ rounds. In each round the player needs to resolve the dilemma of whether to explore an arm which has not been pulled in the past, or whether to exploit the knowledge of the cumulative losses of the arms that have been pulled in the past. We provide a pool based, sequential AL algorithm called LCB-AL, which is motivated by applying algorithmic ideas from the problem of multi-armed bandits to the problem of AL. In order to do so we build a bridge between the MAB problem and AL problem, providing an equivalence between the arms of a MAB problem, and the hypothesis in $\mathcal{H}$, and mitigating the problem of absence of an explicit loss signal in AL, unlike MAB. Establishing this analogy is not very straightforward, but once done allows us to readily use tools such as lower confidence bounds (Auer et al., 2002a), and self-concordant regularization (Abernethy et al., 2008) in the design of LCB-AL. To our knowledge, our work is one of the first in trying to use bandit type ideas for active learning, and we strongly believe that one can build extremely practical, yet very simple and scalable algorithms by understanding the interplay between multi-armed bandits and active learning.

In section 2 we take the first steps towards building an analogy between MAB and AL. This inspires us to use a very successful algorithm from the MAB literature, for the problem of AL. We build the technical tools needed to fill in the details of our proposed algorithm in section 3. Section 4 discusses related work, and section 5 compares LCB-AL with other active learning algorithms on various datasets.

## 2 Towards an analogy between Multi-armed bandits and Active Learning

In active learning the goal is to find a hypothesis $h \in \mathcal{H}$ with low risk, by using as little labeled data as possible. In other words, we want to quickly estimate the risk of different hypothesis, and discard suboptimal hypothesis. In MAB, the goal of the player is to design a strategy, that minimize the cumulative loss suffered by the player over $T$ rounds. If the player knew the arms with the smallest possible cumulative loss then the optimal strategy would be to pull this arm in each and every round. Hence, in MAB the player wants to quickly detect the (near) optimal arm to pull. Looking from the lens of MAB, it is now natural to think of AL problem as a MAB problem, where the arms of the MAB are the different hypothesis in $\mathcal{H}$. While this is a satisfying connection there are two issues that still need to be resolved. 1) In the MAB problem, in each round we pull an arm of the MAB. If arms of the MAB were equivalent to the different hypothesis in $\mathcal{H}$, then how do we decide which "hypothesis to pull". 2) In MAB the player gets to see an explicit loss signal at the end of each round. However, in AL there is no such explicit loss signal, instead the feedback that is received is the label of the queried point $x$. Hence, the next question that arises is how could one use the label information as some kind of a loss signal? The following subsections attempt to resolve these issues.

### 2.1 Which hypothesis to pull?

A very popular approach in MAB to mitigate the exploration-exploitation trade-off is via the use of lower confidence bounds (LCB) (Auer et al., 2002b,a; Audibert et al., 2009a; Bubeck and Cesa-Bianchi, 2012) [3]. In the LCB approach, at the end of round $t$, for each arm $a$ in the set of arms, we build a lower confidence bound, $\text{LCB}_t(a)$ for the cumulative loss the player would have suffered, in hindsight, had he pulled arm $a$ for the first $t$ rounds. The choice of arm $a_{t+1}$ to be pulled in the next round, i.e. round $t+1$ is the solution to the optimization problem $a_{t+1} \in \arg\min \text{LCB}_t(a)$. Such lower confidence bounds can be derived via concentration inequalities (Auer et al., 2002b; Audibert et al., 2009b), and are generally expressed as $LCB_t(a) \stackrel{\text{def}}{=} \hat{L}_t(a) - U(\hat{L}_t(a))$, where $\hat{L}_t(a)$ is an estimate of the cumulative loss of arm $a$, the player would have suffered had he pulled $a$ each time for the first $t$ rounds, and $U(\hat{L}_t(a))$ is some measure of uncertainty (typically variance) of the cumulative

---
[1]Usually in the literature on MAB it is common to talk of rewards. For our purposes, it would be more convenient to talk of losses rather than rewards.

[2]We shall consider the non-stochastic bandit problem

[3]Traditionally it has been called as the upper-confidence bound algorithm. Since, we are dealing with losses and not rewards, it is useful for our purpose, to rename this as LCB

loss of $a$, at the end of round $t$. The reason behind the success of confidence bounds in the MAB problem can be explained by the fact that $\text{LCB}_t(a)$ captures both the knowledge of the cumulative loss, via $\hat{L}_t(a)$, as well as the uncertainty in this estimate, via $U(\hat{L}_t(a_t))$. By pulling the arm $a_{t+1}$ in round $t+1$ of our MAB algorithm, and by updating our estimate of the cumulative loss of arm $a_{t+1}$, our updated estimate $\hat{L}_{t+1}(a_{t+1})$ is a better estimator as $U(\hat{L}_{t+1}(a_{t+1}))$ is potentially smaller than $U(\hat{L}_t(a_{t+1}))$.

One could use a similar technique even in AL. If one had some kind of a LCB on the risk of each hypothesis, then we could equate pulling a hypothesis as solving the optimization problem $h_{t+1} \in \arg\min_{h \in \mathcal{H}} \text{LCB}_t(h)$, where $\text{LCB}_t(h)$ is the lower confidence bound on the risk of $h \in \mathcal{H}$. An LCB for $R(h)$ can be obtained by utilizing the labeled data gathered over the run of the algorithm.

## 2.2 Absence of a loss signal in AL

When an arm is pulled in the MAB setting, the player suffers a loss, and this loss is used to update the LCB of the chosen arm. However, in AL there is no such explicit loss signal. One might come up with a proxy loss signal for the active learning problem which can then be used to update the lower confidence bound of all the hypothesis in $\mathcal{H}$. However, we take a different approach. The utility of the loss signal when the arm $a_t$ is pulled in round $t$ of the MAB problem is two folds. Firstly to update the cumulative loss of $a_t$, and secondly to decrease the uncertainty in the estimate of the cumulative loss of $a_t$. In AL when a certain point $x$ is queried for its label, then this label information can be utilized to improve the error estimate of $h_t$ as well as other hypothesis. Hence, it makes sense to query $\mathcal{O}$, for the label of some point $x$ in $P$, such that its label information maximally reduces the variance of the estimate of risk of $h_t$. Hence, by conceptually viewing label information as a mechanism to reduce the variance of the risk estimate of different hypothesis, we have a disciplined way of deciding which points to query. Table 1 summarizes the analogy between AL and MAB.

## 2.3 Rough outline of LCB-AL.

LCB-AL is a sequential algorithm where in each round a probability distribution is placed on $\mathcal{P}$, and a single point is sampled from $\mathcal{P}$. We additionally assume that the hypothesis class $\mathcal{H}$ is convex. For the sake of simplicity we shall allow re-querying points in the pool, i.e. if a point $x_i$ was first queried in some round $t_1$, then it might once again be queried in round $t_2$. If such a re-querying happens then we use the label that was returned by $\mathcal{O}$ in round $t_1$. [4] In order to construct lower confidence bounds on the risk of $h$ we use importance weighting along with Bernstein type inequalities for martingales. The advantage of using importance weights, is that they facilitate data reuse, when an actively sampled data with one hypothesis class, is used in the future to learn a model from a different hypothesis class. The problem with such importance weighted estimators is that they have very high variance. The seminal work of Abernethy et al. (2008) showed that one could tackle the high variance of the importance weighted estimators, via the use of self-concordant barriers (Nesterov and Nemirovsky, 1994). This inspires us to use the self-concordant barrier of $\mathcal{H}$ along with lower confidence bounds in our algorithm. As a result in each round (see step 14 of algorithm 1) we solve the optimization problem

$$h_{t+1} \in \arg\min_{h \in \mathcal{H}} LCB_t(h) + \mathcal{R}(h),$$

where $\mathcal{R}(h)$ is the self-concordant barrier of $\mathcal{H}$. Using $h_{t+1}$ we induce a sampling distribution over the pool $\mathcal{P}$, at the start of round $t+1$ (see step 4 of algorithm 1). As discussed in section 2.2, the probability distribution is such that it minimizes the (conditional) variance of the estimate of risk of $h_t$. We shall make this clear in section 3.2.

## 3 Risk Estimates and Confidence Bounds

We begin with the notation that will be required to develop our confidence bounds. Let $p_i^t$ be the probability of querying $x_i$ in round $t$, and $Q_i^t \in \{0, 1\}$ be the random variable which takes the value 1 if $x_i$ was queried in round $t$, and 0 otherwise. Hence $\mathbb{E}[Q_i^t | p_i^t] = 1$. For convenience, we shall denote by $Q_{1:n}^{1:t}$ the collection of random variables $Q_1^1, \ldots Q_1^t, \ldots, Q_n^1, \ldots, Q_n^t$. Let $Z_i^t \stackrel{\text{def}}{=} y_i Q_i^t$. Denote by $x_{1:n}$ the collection of random variables $x_1, \ldots, x_n$. Also let $[x]_+ = \max\{x, 0\}$.

We shall make the following independence assumption:

**Assumption 1.** *If $x_i$ has not been queried up until the start of round $t$, then $p_i^t \perp\!\!\!\perp y_i | x_{1:n}, Z_{1:n}^{1:t-1}$.*

For any hypothesis $h \in \mathcal{H}$, define $\hat{L}_t(h) \stackrel{\text{def}}{=} \frac{1}{nt} \sum_{i=1}^n \sum_{\tau=1}^t \frac{Q_i^\tau}{p_i^\tau} L(y_i h(x_i))$. $\hat{L}_t(h)$ is an unbiased estimator of the risk of the hypothesis $h$, and was first proposed by Ganti and Gray (2011).

---

[4] An algorithm similar to the one suggested in this paper, can be designed such that re-querying is not allowed. This requires different type of estimators, and the expressions for $\text{LCB}_t(h)$ turned out to be rather complicated. Hence, for simplicity of exposition we allow re-querying of points in this paper.

| MAB | AL |
|---|---|
| Arms | Hypothesis |
| Loss signal on pulling an arm helps improve cumulative loss estimates | Sampling distribution designed to reduce variance of risk estimates of hypothesis |

Table 1: The analogy between MAB and AL that is used as a guiding principle for the design of LCB-AL

**Algorithm 1** LCB-AL **Input:** $\mathcal{P} = \{x_1, \ldots, x_n\}$, Loss function $L(\cdot)$, Budget $B$, Labeling Oracle $\mathcal{O}$, $p_{\min}$

1: Set $h_1 = 0, t = 1$.
2: **while** num_queried $\leq B$ **do**
3:   **for** $x_i \in \mathcal{P}$ **do**
4:
$$\bar{y}_i = \begin{cases} y_i & \text{if } x_i \in \mathcal{Q}_{t-1} \\ sign(h_t(x_i)) & \text{otherwise} \end{cases} \quad (1)$$
5:     $p_i^t \leftarrow p_{\min} + (1 - np_{\min}) \frac{L(\bar{y}_i h_t(x_i))}{\sum_{x_i \in \mathcal{P}} L(\bar{y}_i h_t(x_i))}$.
6:   **end for**
7:   Sample a point (say $x$) from the probability vector $p^t$.
8:   **if** $x$ was not queried in the past **then**
9:     Query $\mathcal{O}$ for the label $y$ of $x$.
10:     num_queried $\leftarrow$ num_queried $+ 1$
11:   **else**
12:     Reuse the label of $x$.
13:   **end if**
14:   Solve: $h_{t+1} = \arg\min_{h \in \mathcal{H}} \text{LCB}_t(h) + \lambda_t \mathcal{R}(h)$.
15:   $t \leftarrow t + 1$
16: **end while**
17: Return $h_B$.

Finally let $\mathcal{Q}_t \stackrel{\text{def}}{=} \{x_i \in \mathcal{P} | \sum_{\tau=1}^{t} Q_i^\tau > 0\}$. Let $\mathcal{F}_\tau \stackrel{\text{def}}{=} \sigma(x_{1:n}, Z_{1:n}^{1:\tau})$ be the smallest sigma algebra that makes the random variables $x_{1:n}, Z_{1:n}^{1:\tau}$ measurable. Clearly $\mathcal{F}_1 \subset \ldots \subset \mathcal{F}_t$ form a filtration. Also we shall assume that our loss function is a convex function of the margin $yh(x)$, and is upper bounded by $L_{\max} < \infty$ for all $x \in \mathcal{P}, h \in \mathcal{H}$. Popular loss functions such as logistic loss, exponential loss, squared loss all satisfy these criteria.

### 3.1 Constructing lower confidence bounds

Utilizing the unbiased estimator $\hat{L}_t(h)$, along with Bernstein type inequalities for martingales allows us to construct lower confidence bounds for $R(h)$. We shall begin with the standard Azuma-Hoeffding bound for martingale difference sequences.

**Theorem 1.** *[Azuma-Hoeffding inequality] Let $X_1, X_2, \ldots$ be a martingale difference sequence w.r.t a filtration $\mathcal{F}_1 \subset \mathcal{F}_2 \subset \ldots$. If for each $i \geq 1$, $|X_i| \leq c_i$.*

Then,
$$\mathbb{P}[|\sum_{i=1}^{n} X_i| \geq \epsilon] \leq 2\exp\left(-\frac{2\epsilon^2}{\sum_{i=1}^{n} c_i^2}\right)$$

We shall also need the following Bernstein type result from Bartlett et al. (2008).

**Theorem 2.** *Let $M_1, \ldots, M_t$ be a martingale difference sequence (MDS), w.r.t. the filtration $\mathcal{F}_1 \subset \ldots \subset \mathcal{F}_t$, with $|M_\tau| \leq b$. Let $\mathbb{V}_\tau M_\tau \stackrel{\text{def}}{=} \mathbb{V}(M_\tau | \mathcal{F}_{\tau-1})$, and $\sigma^2 \stackrel{\text{def}}{=} \sum_{\tau=1}^{t} \mathbb{V}_\tau M_\tau$. Then we have, for any $\delta < 1/e$, and $t \geq 4$, with probability at least $1 - \delta\log(t)$*

$$\sum_{\tau=1}^{t} M_\tau < 2\max\{2\sigma, b\sqrt{\log(1/\delta)}\}\sqrt{\log(1/\delta)}.$$

**Lemma 1.** *For any fixed $h \in \mathcal{H}$, $t \geq 4, \delta < 1/e$, with probability at least $1 - \delta\log(t)$, we have*

$$\frac{1}{n}\sum_{i=1}^{n}\sum_{\tau=1}^{t} \frac{Q_i^\tau}{p_i^\tau} L(y_i h(x_i)) - \frac{t}{n}\sum_{i=1}^{n} L(y_i h(x_i)) \leq$$

$$2\max\left(\frac{2}{n}\sqrt{\sum_{\tau=1}^{t}\sum_{i=1}^{n}\frac{L^2(y_i h(x_i))}{p_i^\tau} - \left(\sum_{i=1}^{n} L(y_i h(x_i))\right)^2},\right.$$

$$\left. L_{\max}\left(1 + \frac{1}{np_{\min}}\right)\sqrt{\log(1/\delta)}\right)\sqrt{\log(1/\delta)}$$

*Proof.* Let,
$$M_\tau \stackrel{\text{def}}{=} \frac{1}{n}\sum_{i=1}^{n} \frac{Q_i^\tau}{p_i^\tau} L(y_i h(x_i)) - \frac{1}{n}\sum_{i=1}^{n} L(y_i h(x_i)). \quad (2)$$

Utilizing the independence assumption it is easy to see that $\mathbb{E}[M_\tau | \mathcal{F}_{\tau-1}] = 0$. Hence $M_1, \ldots, M_t$ form a martingale difference sequence w.r.t. the filtration $\mathcal{F}_1, \ldots, \mathcal{F}_t$. In order to apply theorem 2 we need estimates for the sum of conditional variances, and the range of $|M_\tau|$. We proceed to establish upper bounds on these quantities now.

From equation 2 and triangle inequality we get

$$|M_\tau| \leq \frac{1}{n}|\sum_{i=1}^{n} \frac{Q_i^\tau}{p_i^\tau} L(y_i h(x_i))| + \frac{1}{n}|\sum_{i=1}^{n} L(y_i h(x_i))|$$

$$\leq L_{\max}\left(1 + \frac{1}{np_{\min}}\right). \quad (3)$$

$$\sigma^2 \stackrel{\text{def}}{=} \sum_{\tau=1}^{t} \mathbb{E}[M_\tau^2 | \mathcal{F}_{\tau-1}]$$

$$= \sum_{\tau=1}^{t} \frac{1}{n^2} \mathbb{E}\Big[\frac{Q_i^\tau}{(p_i^\tau)^2} L^2(y_i h(x_i))$$
$$+ \underbrace{\frac{2}{n} \sum_{i \neq j} \frac{Q_i^\tau Q_j^\tau}{p_i^\tau p_j^\tau} L(y_i h(x_i)) L(y_j h(x_j))}_{=0} \quad (4)$$
$$- \frac{1}{n^2} \Big(\sum_{i=1}^{n} L(y_i h(x_i))\Big)^2 | \mathcal{F}_{\tau-1}\Big]$$

$$= \frac{1}{n^2} \sum_{\tau=1}^{t} \sum_{i=1}^{n} \frac{L^2(y_i h(x_i))}{p_i^\tau} - \frac{1}{n^2} \Big(\sum_{i=1}^{n} L(y_i h(x_i))\Big)^2. \quad (5)$$

In equation 4 we used the fact that in each round only one point is queried. The result now follows by the application of theorem 2 with $b, \sigma^2$ defined as in equations 3, 5 respectively. □

While this theorem enables us to construct lower bounds for the risk of the hypothesis, the major problem is that the RHS of theorem 1 depends on the labels of all the points in the pool and not just the queried labels. We shall now provide an estimator for the variance term $\sigma^2$.

**Lemma 2.** *With probability at least $1 - \delta$, we have*

$$\sigma^2 \leq \frac{1}{n^2} \Big[\sum_{\substack{i=1:n \\ \tau=1:t}} \frac{Q_i^\tau}{(p_i^\tau)^2} L^2(y_i h(x_i)) - \Big(\sum_{\mathcal{Q}_t} L(y_i h(x_i))\Big)^2$$
$$+ \frac{L_{\max}^2 \sqrt{2t \log(1/\delta)(n-1)}}{\sqrt{p_{\min}}}\Big]_+$$

*Proof.* From the proof of theorem 1 we have

$$\sigma^2 = \underbrace{\frac{1}{n^2} \sum_{\tau=1}^{t} \sum_{i=1}^{n} \frac{L^2(y_i h(x_i))}{p_i^\tau}}_{I_1} - \underbrace{\frac{1}{n^2} \Big(\sum_{i=1}^{n} L(y_i h(x_i))\Big)^2}_{I_2}.$$

A simple lower bound on $I_2$ is

$$\hat{I}_2 \stackrel{\text{def}}{=} \frac{1}{n^2} \Big(\sum_{\mathcal{Q}_t} L(y_i h(x_i))\Big)^2. \quad (6)$$

Now let

$$\hat{I}_1 \stackrel{\text{def}}{=} \frac{1}{n^2} \sum_{\tau=1}^{t} \sum_{i=1}^{n} \frac{Q_i^\tau}{(p_i^\tau)^2} L^2(y_i h(x_i)).$$

Define

$$M_\tau \stackrel{\text{def}}{=} \frac{1}{n^2} \sum_{i=1}^{n} \frac{Q_i^\tau}{(p_i^\tau)^2} L^2(y_i h(x_i)) - \frac{1}{n^2} \sum_{i=1}^{n} \frac{1}{p_i^\tau} L^2(y_i h(x_i))$$

Once again utilizing our independence assumption, we conclude that $M_1, \ldots M_t$ form an MDS w.r.t. the filtration $\mathcal{F}_1, \ldots, \mathcal{F}_t$. Applying theorem 1 to this MDS, we get with probability at least $1 - \delta$

$$|\sum_{\tau=1}^{t} M_\tau| \leq \frac{L_{\max}^2 \sqrt{2t \log(1/\delta)}}{n^2} \sqrt{\frac{n-1}{p_{\min}}}. \quad (7)$$

The result follows from equations 6, 7 □

We are now ready to establish a lower confidence bound on the risk of hypotheses in $\mathcal{H}$.

**Theorem 3.** *Let $|\mathcal{H}| < \infty$. With probability at least $1 - |\mathcal{H}|\delta(2 + T \log(T/e))$, for all $h \in \mathcal{H}, 4 \leq t \leq T$, and $\delta < 1/e$, we have*

$$R(h) \geq \Big[\hat{L}_t(h) - \frac{2}{t} \log(1/\delta) L_{\max}\Big(1 + \frac{1}{np_{\min}}\Big)$$
$$- \frac{4}{nt} \sqrt{V_t \log(1/\delta)} - \sqrt{\frac{L_{\max}^2 \log(1/\delta)}{2n}}\Big]_+ \quad (8)$$

*where*

$$V_t \stackrel{\text{def}}{=} \Big[\sum_{\substack{i=1:n \\ \tau=1:t}} \frac{Q_i^\tau}{(p_i^\tau)^2} L^2(y_i h(x_i)) - \Big(\sum_{\mathcal{Q}_t} L(y_i h(x_i))\Big)^2$$
$$+ \frac{L_{\max}^2 \sqrt{2t \log(1/\delta)(n-1)}}{\sqrt{p_{\min}}}\Big]_+ \quad (9)$$

*Proof.* For any fixed $h \in \mathcal{H}$, and $t \leq T$, we have from theorems 1, 2, Hoeffding inequality, and the union bound that with probability at least $1 - \delta \log(t) - 2\delta$

$$R(h) \geq \Big[\hat{L}_t(h) - \frac{2}{t} \log(1/\delta) L_{\max}(1 + \frac{1}{np_{\min}})$$
$$- \frac{4}{nt} \sqrt{V_t \log(1/\delta)} - \sqrt{\frac{L_{\max}^2 \log(1/\delta)}{2n}}\Big]_+. \quad (10)$$

Applying union bound over all hypothesis and over all $t \geq 4$, and approximating $n-1$ with $n$ we get the desired result. □

**Specification of $\text{LCB}_t(h)$.** Theorem 3 provides us with an expression for $\text{LCB}_t(h)$. For the purpose of solving the optimization in step 7 of our LCB-AL algorithm, we can set

$$\text{LCB}_t(h) \stackrel{\text{def}}{=} \hat{L}_t(h) - \frac{4}{nt} \sqrt{\log(1/\delta) V_t}, \quad (11)$$

where $V_t$ is shown in equation 9.

## 3.2 Query probability distribution in each round of LCB-AL

The only thing that is left to be motivated in LCB-AL is the choice of probability distribution in step 3. As explained in section 2.2 we want to use a sampling distribution, such that the conditional variance of the risk estimate $\hat{L}_t(h)$ of $h_t$ is minimized. We shall now show how the sampling distribution should be designed in order to achieve this goal. Let $\Delta \subset \mathbb{R}_+^n$ be the probability simplex. Let $V_t(\cdot)$ denote the variance, conditioned on $x_{1:n}, Z_{1:n}^{1:t-1}$. Let $p^t \stackrel{\text{def}}{=} (p_1^t, \ldots, p_n^t) \in \Delta$. At the start of round $t$, the desired sampling distribution, $p^t$ should satisfy

$$p^t = \arg\min_{\tilde{p}^t \in \Delta} \mathbb{V}_t \Big[ \underbrace{\frac{1}{nt} \sum_{\substack{i=1:n \\ \tau=1:t}} \frac{\tilde{Q}_i^\tau}{\tilde{p}_i^\tau} L(y_i h_t(x_i))}_{\hat{L}_t(h_t)} \Big]$$

$$= \arg\min_{\tilde{p}^t \in \Delta} \sum_{i=1}^n \frac{\mathbb{E}_t L^2(y_i h_t(x_i))}{\tilde{p}_i^\tau}$$

Solving the above optimization problem yields the simple solution $p_i^t \propto \sqrt{\mathbb{E}_t L^2(y_i h_t(x_i))}$. If $x_i \in \mathcal{Q}_{t-1}$, then the label $y_i$ is known and hence, we let $p_i^t \propto L(y_i h_t(x_i))$. If $x_i \notin \mathcal{Q}_{t-1}$, then since $y_i$ is yet unknown, we let $p_i^t \propto L(|h_t(x_i)|)$. This is equivalent to taking $y_i$ to be equal to $\text{sgn}(h_t(x_i))$ (see steps 3, 4 of algorithm 1). This scheme encourages querying points which have small margin w.r.t the current classifier, $h_t$, or points which have already been queried for their label, but on which the current hypothesis, $h_t$ suffers a large loss. In any round, the minimum probability of querying any point is $p_{\min}$. This guarantees that $\hat{L}_t(h)$ is an unbiased estimator of risk of $h$.

## 4 Related Work

A variety of pool based AL algorithms have been proposed in the literature employing various query strategies. Some of the popular querying strategies include uncertainty sampling, where the active learner queries the point whose label it is most uncertain about (Lewis and Gale, 1994; Settles and Craven, 2008; Tong and Chang, 2001). Usually the uncertainty in the label is calculated using certain information-theoretic criteria such as entropy, or variance of the label distribution. Seung et al. (1992) introduced the query-by-committee (QBC) framework where a committee of potential models, which all agree on the currently labeled data is maintained and, the point where most committee members disagree is considered for querying. Other frameworks include querying the point, which causes the maximum expected reduction in error (Zhu et al., 2003; Guo and Greiner, 2007), variance reducing query strategies such as the ones based on optimal design (Flaherty et al., 2005; Zhang and Oles, 2000). A very thorough literature survey of different active learning algorithms has been done by Settles (2009). AL algorithms that are consistent and have provable label complexity have been proposed for the agnostic setting for the 0-1 loss in recent years (Dasgupta et al., 2007; Balcan et al., 2009). Hanneke and Yang (2012) recently provided a disagreement region based algorithm, with provable guarantees for active learning with general loss functions.

Algorithmically, LCB-AL is similar in flavor to the UPAL algorithm introduced by Ganti and Gray (2011). In the UPAL algorithm the authors suggested minimizing an unbiased estimator of risk of $h$, and a sampling distribution that was in proportion to the entropy of the prediction on the pool. However as we suggested the use of self-concordant regularizer is very crucial in tackling the high variance of our estimators. As we show in our experiments (see section 5) the use of self-concordant barrier as a regularizer, helps lend stability to our algorithm, and consequently LCB-AL performs better than UPAL.

To our knowledge there has been only one other paper bridging the world of active learning and MAB. Baram et al. (2004) proposed a meta-active learning algorithm called COMB. COMB was an implementation of the EXP4 algorithm for MAB with expert advice, where the different active learning algorithms are the various "experts" and the different points in the pool are the arms of the MAB. Briefly, in each round, each of the experts suggest a sampling distribution on the pool. COMB maintains an estimate of the error rate of each expert, and uses exponential weighting to come up with a sampling distribution on the pool. In order to estimate the error rate of each of the experts, the authors proposed a proxy reward function of querying a point in terms of the entropy of label distribution of the unlabeled pool, induced by the classifier obtained on the labeled dataset gathered by COMB till the current iteration. In a way, the concept of reward seems inevitable in their formulation because the unlabeled points in the pool are treated as arms of the MAB. In contrast, we think of the arms of the bandit as the different hypothesis, and querying a data point, as the process of improving our estimate of the risk of the different hypothesis. Hence, we bypass the need for an explicit reward signal, yet utilize MAB ideas for AL.

## 5 Experiments

We implemented LCB-AL in MATLAB, and compared it with some previously proposed active learning algorithms on four UCI datasets. The competing algo-rithms are UPAL (Ganti and Gray, 2011), BMAL (Hoi et al., 2006), and a passive learning (PL) algorithm that minimizes the regularized logistic loss. UPAL is a sequential, pool based algorithm that minimizes the unbiased estimator $\hat{L}_t(h)$, of the risk of a hypothesis $h$, along with the squared $L_2$ norm of $h$ in each round. BMAL is a batch mode active learning algorithm introduced by Hoi et al. (2006). Hoi et al. in their paper showed superior empirical performance of BMAL over other pool based active learning algorithms, and this is the primary motivation for using BMAL in our experiments. Given a pool $\mathcal{P}$, and a budget $B$, BMAL chooses a set of $B$ points that minimize the Fisher information ratio between the set of unqueried and the queried points. The authors then propose a monotonic submodular approximation to the original Fisher ratio objective, which is optimized by a greedy algorithm. In order to keep our experiments simple, and to facilitate easy comparison, we restricted our hypothesis class to the set of linear hypothesis of bounded $L_2$ norm $\mathcal{H} = \{h : ||h|| \leq R\}$. For this set $\mathcal{H}$, the self-concordant regularizer $\mathcal{R}(h)$ is equal to $-\log(R^2 - ||h||^2)$ (Abernethy et al., 2008), where $R > 0$ was provided as an input to LCB-AL. For our implementation of LCB-AL, we used a slightly different definition for $\text{LCB}_t(h)$, than the one proposed in equation 11. Let

$$\text{LCB}'_t(h) \stackrel{\text{def}}{=} \hat{L}_t(h) - C_t\sqrt{V'_t} - \lambda_t \log(R^2 - ||h||^2), \quad (12)$$

where

$$V'_t \stackrel{\text{def}}{=} \Big[\sum_{\substack{i=1:n \\ \tau=1:t}} \frac{Q_i^\tau}{(p_i^\tau)^2} L^2(y_i h(x_i)) - \Big(\sum_{\mathcal{Q}_t} L(y_i h(x_i))\Big)^2\Big]_+.$$

The definition of $\text{LCB}'_t(h)$ is almost similar to the one suggested by equation 11 except that the terms that were independent of $h$, in $\text{LCB}_t(h)$ were dropped to get $\text{LCB}'_t(h)$. $C_t$, and $\lambda_t$ in all our experiments were set to $\frac{\sqrt{\log(t)}}{10}$, and $\frac{100nt}{\left(\sum_{i=1}^n \sum_{\tau=1}^{t-1} \frac{Q_i^\tau}{p_i^\tau}\right)^{1/3}}$ respectively. We used minFunc [5] to solve all of our optimization problems. We used a budget of 300 points. Finally, each of the dataset was scaled to $[-1, 1]^d$. Since, UPAL and LCB-AL are randomized algorithms, on each dataset, we ran them 10 times each, and report averaged measurements.

### 5.1 Experimental comparisons of different algorithms

Figure 1 shows the test error of the hypothesis obtained, corresponding to the number of unique queries

---

[5]minFunc can be downloaded from http://www.di.ens.fr/~mschmidt/Software/minFunc.html

---

made to the oracle, by each algorithm. Table 2 shows the error rate on the test set, of each algorithm, once the budget is exhausted. On three of the datasets, namely MNIST, Abalone, and Statlog utilizing an active learner is better than a passive learner. On MNIST, the performance of LCB-AL and UPAL are nearly equal as far as the final test error goes, and both are better than BMAL. On abalone LCB-AL is better than both BMAL and UPAL, while on Statlog the final error achieved by BMAL is better than UPAL, and also LCB-AL, though the difference between LCB-AL and BMAL is pretty narrow. On Whitewine, passive learner is better than any of the active learners. In order to gain an insight into how well each of the learning algorithm learns with each query to the oracle, we also report the cumulative error rate of each algorithm, summed over all the queries. The cumulative error rate of a learning algorithm is nothing but the area under the curve (AUC) of error-rate vs number of queries to the oracle. Even on this measure, LCB-AL and UPAL are better than BMAL on MNIST, Abalone and Whitewine datasets. On Abalone, and Statlog the AUC of LCB-AL is appreciably smaller than that of UPAL.

### 5.2 Comparing UPAL with LCB-AL

From our first set of experiments it looks like UPAL is just as good as LCB-AL if not any better. e.g. on the MNIST dataset, there is almost no difference between LCB-AL and UPAL. Since, both LCB-AL and UPAL are randomized algorithms it makes sense to measure the fluctuations in the performance of both the algorithms. Table 3 gives the standard deviation of AUC over all the runs for both LCB-AL and UPAL. It is clear that the standard deviation of AUC for LCB-AL is uniformly smaller than that of UPAL over all datasets, and the difference in the standard deviations is largest for the MNIST dataset. This can be explained by the fact that, the unbiased estimator of risk used in UPAL is a high variance estimator, and hence not a reliable estimator of the risk of a hypothesis. In LCB-AL, by utilizing lower confidence bounds, and the self-concordant regularizer, we are able to tackle the high variance of our estimator, and at the same time harness the variance for exploration in the hypothesis space. In fact, a similar phenomenon occurs even in the MAB setting, where algorithms built only on unbiased estimators, such as EXP3 (Auer et al., 2002b), achieve optimal performance only on an average, whereas algorithm using confidence bounds such as EXP3.P achieve optimal performance with high probability.

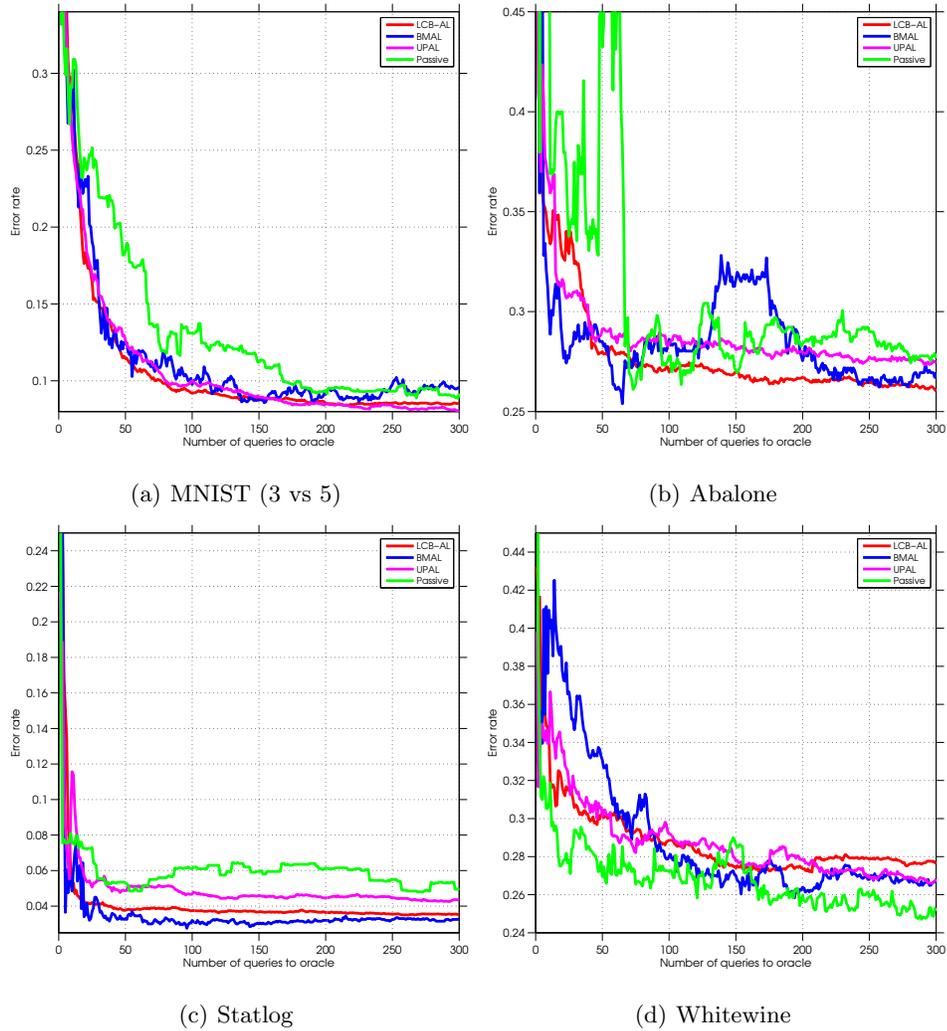

Figure 1: Error rate of different learning algorithms with the number of queries made to the oracle.

| Dataset | LCB-AL | | UPAL | | BMAL | | PL | |
|---|---|---|---|---|---|---|---|---|
| MNIST | 0.0808 | 33.27 | 0.0809 | 32.75 | 0.0958 | 34.89 | 0.0918 | 40.08 |
| Abalone | 0.2604 | 83.49 | 0.2747 | 86.60 | 0.2695 | 86.21 | 0.2766 | 93.60 |
| Statlog | 0.0354 | 12.59 | 0.0433 | 14.97 | 0.0330 | 11.33 | 0.05 | 18.06 |
| Whitewine | 0.2771 | 86.30 | 0.2682 | 86.21 | 0.2665 | 86.95 | 0.2517 | 80.94 |

Table 2: Comparison of various active learning algorithms and passive learner on various datasets. In this table we report both the error rate of each learner after it has exhausted its budget, as well as the area under the curve of error rate vs number of queries made for each learning algorithm.

| Dataset | LCB-AL | UPAL |
|---|---|---|
| MNIST | 3.8604 | 5.0132 |
| Abalone | 2.6512 | 2.6869 |
| Statlog | 0.7944 | 1.6691 |
| Whitewine | 2.4097 | 2.9992 |

Table 3: Standard deviation of the AUC of LCB-AL and UPAL on different datasets.

# 6  Discussion

We proposed LCB-AL a multi-armed bandit inspired pool based active learning algorithm. By viewing the problem of active learning as quickly detecting the hypothesis with (near) optimal risk, we view the problem of active learning as similar to a MAB problem with the arms being the different hypothesis. By building lower confidence bounds on the risk of each hypothesis we are able to perform exploration in the hypothesis space. By conceptually investigating the role of a loss signal in MAB, we are able to design a sampling distribution from which we sample the points to be queried. Experimental results suggest that our algorithm is both more accurate, and also more stabler than competing active learning algorithms.

In the near future we would like to investigate LCB-AL theoretically. Two properties of LCB-AL are worth investigating. Firstly, can we guarantee that the excess risk of our algorithm goes to 0, as $n \to \infty, B \to \infty$? Secondly, what is the budget $B$, required in order to guarantee an excess risk of $\epsilon$?

An immediate extension of this work could be to investigate how different concentration inequalities can be utilized to give different lower confidence bounds for the risk of a hypothesis. This has proven to be an attractive idea in the MAB setting and it is generally accepted that tighter concentration inequalities lead to better algorithms for MAB (Audibert et al., 2009a; Salomon and Audibert, 2011). We would expect something similar to happen even in AL.

On a more high level we believe that there is tremendous potential for ideas from multi-armed bandits, and various other extensions of multi-armed bandits such as contextual bandits, bandit optimization, to be used for active learning problems. Most of these algorithms are very simple, efficient and hence should be useful in designing simple, efficient active learning algorithms.

## References

J. Abernethy, E. Hazan, and A. Rakhlin. Competing in the dark: An efficient algorithm for bandit linear optimization. In *Proceedings of the 21st Annual Conference on Learning Theory (COLT)*, volume 3, page 3, 2008.

Jean-Yves Audibert, Rémi Munos, and Csaba Szepesvári. Exploration-exploitation tradeoff using variance estimates in multi-armed bandits. *Theor. Comput. Sci.*, 2009a.

J.Y. Audibert, R. Munos, and C. Szepesvári. Exploration–exploitation tradeoff using variance estimates in multi-armed bandits. *Theoretical Computer Science*, 410(19):1876–1902, 2009b.

P. Auer, N. Cesa-Bianchi, and P. Fischer. Finite-time analysis of the multiarmed bandit problem. *Machine learning*, 47(2):235–256, 2002a.

P. Auer, N. Cesa-Bianchi, Y. Freund, and R.E. Schapire. The nonstochastic multiarmed bandit problem. *SIAM Journal on Computing*, 32(1):48–77, 2002b.

M.F. Balcan, A. Beygelzimer, and J. Langford. Agnostic active learning. *JCSS*, 75(1), 2009.

Y. Baram, R. El-Yaniv, and K. Luz. Online choice of active learning algorithms. *The Journal of Machine Learning Research*, 5:255–291, 2004.

P.L. Bartlett, V. Dani, T. Hayes, S. Kakade, A. Rakhlin, and A. Tewari. High-probability regret bounds for bandit online linear optimization. *COLT*, 2008.

E.B. Baum and K. Lang. Query learning can work poorly when a human oracle is used. In *IJCNN*, 1992.

Sébastien Bubeck and Nicolò Cesa-Bianchi. Regret analysis of stochastic and nonstochastic multi-armed bandit problems. *arXiv preprint arXiv:1204.5721*, 2012.

W. Chu, M. Zinkevich, L. Li, A. Thomas, and B. Tseng. Unbiased online active learning in data streams. In *SIGKDD*, 2011.

D. Cohn, L. Atlas, and R. Ladner. Improving generalization with active learning. *Machine Learning*, 15(2), 1994.

S. Dasgupta, D. Hsu, and C. Monteleoni. A general agnostic active learning algorithm. *NIPS*, 2007.

Patrick Flaherty, Michael I. Jordan, and Adam P. Arkin. Robust design of biological experiments. In *Neural Information Processing Systems*, 2005.

R. Ganti and A. Gray. Upal: Unbiased pool based active learning. *Arxiv preprint arXiv:1111.1784*, 2011.

Y. Guo and R. Greiner. Optimistic active learning using mutual information. In *IJCAI*, 2007.

S. Hanneke and L. Yang. Surrogate losses in passive and active learning. *arXiv preprint arXiv:1207.3772*, 2012.

S.C.H. Hoi, R. Jin, J. Zhu, and M.R. Lyu. Batch mode active learning and its application to medical image classification. In *ICML*, 2006.

D.D. Lewis and W.A. Gale. A sequential algorithm for training text classifiers. In *SIGIR*, 1994.

Yurii Nesterov and A Nemirovsky. Interior point polynomial methods in convex programming, 1994.

Antoine Salomon and Jean-Yves Audibert. Deviations of stochastic bandit regret. In *Algorithmic Learning Theory*, pages 159–173. Springer, 2011.

B. Settles and M. Craven. An analysis of active learning strategies for sequence labeling tasks. In *EMNLP*, 2008.

Burr Settles. Active learning literature survey. Computer Sciences Technical Report 1648, University of Wisconsin–Madison, 2009.

H.S. Seung, M. Opper, and H. Sompolinsky. Query by committee. In *COLT*, pages 287–294. ACM, 1992.

S. Tong and E. Chang. Support vector machine active learning for image retrieval. In *Proceedings of the ninth ACM international conference on Multimedia*, 2001.

T. Zhang and F. Oles. The value of unlabeled data for classification problems. In *ICML*, 2000.

Xiaojin Zhu, John Lafferty, and Zoubin Ghahramani. Combining active learning and semi-supervised learning using gaussian fields and harmonic functions. In *ICML*, 2003.